\documentclass[11pt]{article}

\usepackage{acl}
\usepackage{algorithm}

\usepackage{algorithmic}
\usepackage{times}
\usepackage{latexsym}
\usepackage[T1]{fontenc}
\usepackage[utf8]{inputenc}
\usepackage{microtype}
\usepackage{inconsolata}
\usepackage{enumitem}
\usepackage{graphicx}
\usepackage{amsmath}
\usepackage{amssymb}
\usepackage{booktabs}
\usepackage{multirow}

\title{Risk-aware Selective Prompting for Hallucination Mitigation in Large Vision-Language Models}

\author{
Yuang Huang\textsuperscript{1} \and
YaFeng Zhang\textsuperscript{2} \and
Zilan Yu\textsuperscript{3} \\
\textsuperscript{1}Shanghai Jiao Tong University \quad
\textsuperscript{2}iFLYTEK \quad
\textsuperscript{3}Tsinghua University \\
\texttt{yuang.huang@sjtu.edu.cn} \quad
\texttt{yfzhang40@iflytek.com} \quad
\texttt{yuzl22@mails.tsinghua.edu.cn}
}

\begin{document}
\maketitle
\begin{abstract}
Prompt-based verification is widely used to mitigate hallucinations in large vision-language models (LVLMs), yet when it helps remains poorly understood. We systematically study verification prompting across two representative LVLM architectures and hallucination benchmarks, and find
that it is a \emph{risk-bearing intervention}: its corrections increase with input difficulty, while newly introduced errors persist across difficulty levels. As a result, always-on prompting helps on hard inputs but offers little benefit---and can harm---easier ones. Our analysis further shows that this behavior is associated with a conservative output shift. Verification prompts redistribute attention from visual tokens
toward instruction tokens and induce a distinct middle-layer entropy pattern absent in a neutral-prompt control, suggesting
instruction-conditioned attention redistribution rather than uniformly improved visual grounding. Motivated by this input-dependent risk, we propose Risk-aware Selective Prompting (RSP), a training-free approach that uses pre-generation uncertainty signals to trigger verification selectively. RSP mitigates the degradation of always-on prompting while
preserving baseline performance, and reveals that effective selection signals vary across architectures.
\end{abstract}

\section{Introduction}

Large vision-language models (LVLMs) such as
LLaVA~\cite{liu2023llava}, InstructBLIP~\cite{dai2023instructblip},
and GPT-4V~\cite{openai2023gpt4v} have achieved remarkable performance
on visual understanding tasks, yet they frequently generate descriptions
of objects absent from the image---a phenomenon known as
\emph{object hallucination}~\cite{li2023pope,rohrbach2018chair}.
Among mitigation strategies, prompt-based verification, such as
instructing the model to check visual evidence or answer more carefully,
is attractive because it requires no retraining or architectural
modification.

In current practice, such verification prompts are often applied
indiscriminately to all inputs (always-on prompting). However, this
raises a question that has received little systematic attention:
\textbf{does verification prompting always help, or can it sometimes
cause harm?} We argue that verification prompting should be viewed not
as a universally beneficial add-on, but as a risk-bearing intervention
that should be applied selectively depending on the input. Recent work
has shown that chain-of-thought prompting can reduce performance on
certain tasks~\citep{sprague2024cot}, but when such harm occurs in
LVLMs, why it occurs, and how to avoid it remain underexplored.

We investigate this question empirically through evaluation on POPE
across two architecturally distinct LVLMs. Our analysis reveals three
findings.

It introduces a relatively stable number of new errors
(\emph{breaks}) across difficulty levels, while the number of errors it
corrects (\emph{fixes}) increases with difficulty. On easy inputs, fixes
and breaks nearly cancel; on hard inputs, fixes far exceed breaks. This
asymmetry explains why always-on prompting helps on adversarial inputs
but provides little benefit and can harm easier ones.

By probing attention maps across transformer layers under verification, neutral, and no-prompt conditions, we find that verification prompts draw attention toward instruction tokens, reduce relative attention to visual tokens, and induce a distinct middle-layer entropy pattern not reproduced by a neutral-prompt control. These observations suggest that verification prompting is associated with instruction-conditioned attention redistribution, rather than clearly indicating uniformly improved visual grounding.

Because prompting is beneficial only for a subset of inputs, we examine
whether internal model states before generation can guide the decision to
prompt. We find that useful signals appear architecture-sensitive:
middle-layer attention entropy is informative for LLaVA-1.5, where visual
tokens are directly exposed to the language model, while output-logit
confidence is more reliable for InstructBLIP, where visual information is
compressed by a Q-Former.

Based on these findings, we propose \textbf{Risk-aware Selective
Prompting (RSP)}, a training-free strategy that triggers verification
only when a pre-generation uncertainty signal indicates high risk. RSP
invokes prompting on only a small fraction of inputs, mitigates the
degradation of always-on prompting on easier inputs, and remains close to
or above baseline performance across our discriminative evaluation.





The contributions of this work are summarized as follows:

\noindent(1)~We identify verification prompting in LVLMs as an
input-dependent, risk-bearing intervention: corrected errors increase
with input difficulty, while newly introduced errors remain relatively
stable across difficulty levels.

\noindent(2)~Through layer-wise attention analysis, we show that this
conservative output shift is accompanied by attention redistribution
from visual tokens toward instruction tokens and a distinct middle-layer
entropy pattern absent in a neutral-prompt control.

\noindent(3)~Motivated by these findings, we propose RSP, a
training-free selective prompting strategy that uses pre-generation
uncertainty signals to trigger verification on inputs identified as
high-risk. We further show that effective routing signals vary across
the two architectures we study.

\section{Related Work}

\subsection{Object Hallucination in LVLMs}

LVLMs often mention objects that are absent from the image, even when
they produce otherwise fluent and plausible responses. This object
hallucination problem is commonly studied with
CHAIR~\citep{rohrbach2018chair}, which measures hallucinated object
mentions in open-ended captions, and POPE~\citep{li2023pope}, which
casts hallucination detection as yes/no object-presence questions under
different difficulty settings. Existing studies link these errors to
object co-occurrence priors~\citep{li2023pope}, weak visual
grounding~\citep{zhou2024lure}, and attention patterns that rely too
heavily on textual context instead of image
tokens~\citep{huang2024opera}. These findings motivate
grounding-oriented interventions, but leave open \emph{when} such
interventions should be applied.

\subsection{Hallucination Mitigation Approaches}

To mitigate object hallucination, prior work has explored interventions
at different stages of the generation pipeline. \emph{Inference-time}
methods intervene during decoding or internal representation processing,
including attention-based penalties~\citep{huang2024opera}, contrastive
decoding~\citep{leng2024vcd,chuang2024dola}, self-correction feedback,
attention-head masking, and activation-space
projection~\citep{zhang2025degf,deng2025maskcd,yang2025nullu,wan2025only}.
These methods are often effective but typically require modified decoding
procedures or per-token intervention during generation.

\emph{Post-hoc} methods verify or revise outputs after generation.
Examples include Woodpecker~\citep{yin2024woodpecker} and
Self-Refine~\citep{madaan2023selfrefine}, which add checking or
refinement stages at inference time.

\emph{Prompt-based} methods offer a lighter alternative. They modify
the input instruction, for example by asking the model to verify its
own response~\citep{dhuliawala2023cove} or to ground answers in visual
descriptions before decoding~\citep{sreyan2025vdgd}. These methods aim
to reduce hallucination without changing the model or decoding
procedure. However, prior evaluations of prompt-based mitigation are
largely aggregate, offering limited analysis of which inputs benefit,
which are harmed, and whether such interventions should be applied
uniformly. This leaves open whether verification prompts should be
treated as universally beneficial aids or as interventions with
input-dependent risk. Our work is complementary to existing mitigation
mechanisms: rather than designing another intervention, we study
\emph{when} prompt-based verification should be applied and show that
internal uncertainty signals enable risk-aware selective prompting.
Our goal is therefore not to outperform specialized mitigation systems,
but to characterize verification prompting as a selective intervention
problem.

\subsection{Uncertainty Estimation and Internal Signals}

Uncertainty estimation in large language models has been studied using
output-based signals such as predictive entropy, semantic entropy, and
sampling-based
agreement~\citep{kuhn2023semantic,kossen2024semantic,nikitin2024kernel}.
These approaches typically operate during or after generation, relying
on token-level probabilities, multiple samples, or completed outputs. A
complementary line of work examines internal model states directly:
probing and intervention studies find that specific layers often play an
important role in distinguishing or mitigating hallucinated
outputs~\citep{chen2024inside,ji2024llminternal,kim2025detecting},
while layer-wise attention analyses show that visual-token attention
varies substantially across depth~\citep{zhang2024seeing}.

To our knowledge, these signals have not been systematically studied as \emph{pre-generation} criteria for prompt-level selective intervention in LVLMs. Relatedly, recent work uses uncertainty for \emph{routing} between different models or inference strategies~\citep{chuang2025confident}, but operates at the model-selection level. In contrast, we study prompt-level selective intervention within a single LVLM. Importantly, we find that the effective pre-generation signal appears \emph{architecture-sensitive}: internal attention entropy is effective for models with direct image-token input (LLaVA-1.5), while output-logit confidence is more predictive for architectures with intermediate compression modules (InstructBLIP's Q-Former), where compressed query tokens make attention-based signals less informative in our experiments. This highlights the need for architecture-aware signal selection.

\section{Method}

We decompose verification prompting into three analytical layers:
(1)~characterizing its behavioral effect, (2)~probing its internal
mechanism, and (3)~designing a selective triggering strategy.

\subsection{Verification Prompting as a Risk-Bearing Intervention}

Given an LVLM $\mathcal{M}$ and an input $(x_\text{img}, x_\text{q})$,
let $y_b = \mathcal{M}(x_\text{img}, x_\text{q})$ denote the baseline
response and $y_p = \mathcal{M}(x_\text{img}, [s; x_\text{q}])$ denote
the response under a verification prompt~$s$. Let $c(y; y^*) \in \{0,1\}$
denote a task-specific correctness indicator. We define four mutually
exclusive outcome categories:
\begin{itemize}[nosep,leftmargin=1.2em]
\item \textbf{Fix}: $c(y_b; y^*)=0$ and $c(y_p; y^*)=1$;
\item \textbf{Break}: $c(y_b; y^*)=1$ and $c(y_p; y^*)=0$;
\item \textbf{Unchanged-correct}: both correct;
\item \textbf{Unchanged-wrong}: both incorrect.
\end{itemize}

As a sample-level diagnostic, we define the net correction count
$|\mathrm{Fix}| - |\mathrm{Break}|$. This decomposition treats
verification prompting as a risk-bearing intervention whose benefit
is not guaranteed for any given input.

\subsection{Probing Attention Redistribution}
\label{sec:probing}

To analyze how verification prompting changes internal computation,
we extract attention maps from all $L$ layers during the prefill pass
before generation begins. For each layer~$l$, we use the final prefill
position and compute the following quantities, averaged over heads.

\paragraph{Attention entropy.}
\begin{equation}
H^{(l)} = -\sum_{i=1}^{n} a_i^{(l)} \log a_i^{(l)}
\end{equation}
where $a_i^{(l)}$ are attention weights over $n$ key positions.
Higher entropy indicates a more dispersed attention distribution.

\paragraph{Visual and instruction attention mass.}
Let $\mathcal{V}$ and $\mathcal{I}$ denote visual-token and
instruction-token positions respectively:
\begin{equation}
M_\text{vis}^{(l)} = \sum_{i \in \mathcal{V}} a_i^{(l)}, \quad
M_\text{inst}^{(l)} = \sum_{i \in \mathcal{I}} a_i^{(l)}.
\end{equation}
A decrease in $M_\text{vis}$ indicates reduced relative attention to
visual tokens; an increase in $M_\text{inst}$ indicates that
instruction tokens absorb a larger share.

To assess whether observed changes are merely due to adding extra text,
we compare three conditions: no prompt (baseline), verification prompt,
and a neutral prompt that adds non-verification text. Full prompt texts
are reported in Appendix~\ref{app:prompts}.

\subsection{Risk-aware Selective Prompting (RSP)}
\label{sec:rsp}

Since verification prompting can both fix and break predictions, we
trigger it selectively. RSP operates as follows:

\paragraph{Signal extraction.}
For each input, a probe prefill pass (without generation) extracts a
scalar uncertainty signal $u(x)$. For LLaVA-1.5, we use layer-$l^*$
attention entropy; for InstructBLIP, we use inverse
first-token confidence, $u(x) = 1 - p_{\mathrm{top1}}$, computed from
the output logit distribution.

\paragraph{Threshold selection.}
On a held-out development set, we sweep candidate trigger rates and
select threshold $\tau$ that maximizes downstream validation
performance.

\paragraph{Selective triggering.}
At inference, the routing rule is $r(x) = \mathbf{1}[u(x) > \tau]$.
If triggered, the model answers with the verification prompt; otherwise,
it uses the baseline prompt.

RSP is training-free: it requires no parameter updates. A probe prefill
is performed for every input; full verification generation is performed
only for the triggered subset.

\section{Experiments}
\label{sec:exp_main}

\subsection{Experimental Setup}
\label{sec:exp_setup}

\paragraph{Models.}
We evaluate two architecturally distinct LVLMs:
(1)~\textbf{LLaVA-1.5-7B}~\cite{liu2023llava}, which concatenates 576
image tokens directly into the language model input (Vicuna-7B backbone,
32 layers); and (2)~\textbf{InstructBLIP-Vicuna-7B}~\cite{dai2023instructblip},
which compresses image features into 32 query tokens via a Q-Former module
before passing them to a Vicuna-7B language backbone.

\paragraph{Benchmarks.}
We use \textbf{POPE}~\cite{li2023pope} for discriminative evaluation,
which probes object hallucination through binary yes/no questions at
three difficulty levels: \emph{random} ($n$\,=\,2{,}910),
\emph{popular} ($n$\,=\,3{,}000), and \emph{adversarial}
($n$\,=\,3{,}000). Images are drawn from MSCOCO 2014 validation.
We additionally evaluate on \textbf{CHAIR}~\cite{rohrbach2018chair}
to probe boundaries on open-ended generation; details are reported
in Appendix~\ref{app:chair}.

\paragraph{Development and evaluation protocol.}
We split POPE-random ($n$\,=\,2{,}910) into a development subset
($n_{\mathrm{dev}}$\,=\,910) and a disjoint test subset
($n_{\mathrm{test}}$\,=\,2{,}000). The development subset is used for
all hyperparameter selection; the test subset and full POPE-popular and
POPE-adversarial splits are used only for final evaluation.

For LLaVA-1.5, we select layer-23 attention entropy as the uncertainty
signal with threshold $\tau$\,=\,1.82 (top~6\% on dev; realized trigger
rate 5--7\% across test splits). For InstructBLIP, attention entropy is near chance
(AUROC\,$\approx$\,0.5), likely due to the compressed visual
representation from the Q-Former; we use inverse first-token
confidence ($1 - p_{\mathrm{top1}}$) as the uncertainty signal,
paired with a short cautious prompt (``Be careful.''), yielding
realized trigger rates of 10--14\% across POPE splits.

\paragraph{Threshold calibration.}
For each model, the routing threshold $\tau$ is selected on the
POPE-random development set ($n_{\mathrm{dev}}$\,=\,910) and then held fixed for all final evaluations within the same model. We do not claim zero-shot threshold transfer across datasets or architectures; RSP requires a modest calibration set but no gradient-based tuning or auxiliary model training.

\paragraph{Computational overhead.}
RSP is training-free but not cost-free: it requires one additional
probe prefill pass for each input to extract the uncertainty signal,
adding moderate wall-clock overhead. Verification generation is then
performed only for the triggered subset of inputs (5--7\% for
LLaVA-1.5, 10--14\% for InstructBLIP). Thus, ``training-free''
refers to the absence of parameter updates or auxiliary model
training, not to zero inference overhead.

\paragraph{Compared methods.}
(1)~\emph{Baseline}: standard inference without verification;
(2)~\emph{Always-on}: verification prompting applied to all inputs;
(3)~\emph{RSP}: our proposed selective prompting.

\paragraph{Generation and metrics.}
All POPE experiments use greedy decoding with
\texttt{max\_new\_tokens}\,=\,10. We report F1 (``yes'' as positive
class), Precision, Recall, and Accuracy. Signal quality is measured by
AUROC. Statistical significance is assessed with paired bootstrap
resampling ($N$\,=\,2{,}000, 95\% CI).

\subsection{Conservative Shift and Difficulty Dependence}
\label{sec:exp_fixbreak}

We first examine the sample-level effects of always-on verification
prompting on LLaVA-1.5. For this diagnostic analysis, we use the full
POPE splits to characterize the aggregate effect of always-on prompting;
all RSP evaluations follow the held-out protocol described in
Section~\ref{sec:exp_setup}.

\begin{table}[t]
\centering
\small
\begin{tabular}{lccccc}
\toprule
\textbf{Split} & \textbf{Fixes} & \textbf{Breaks} & \textbf{Net} & \textbf{$\Delta$Yes\%} & \textbf{$\Delta$F1} \\
\midrule
Random      & 88  & 89  & $-1$   & $-6.4$ & $-0.007$ \\
Popular     & 113 & 91  & $+22$  & $-5.4$ & $-0.002$ \\
Adversarial & 198 & 92  & $+106$ & $-8.8$ & $+0.018$ \\
\bottomrule
\end{tabular}
\caption{Fix/break analysis of always-on verification prompting
(LLaVA-1.5, POPE, full splits). Breaks remain stable ($\sim$90) while
fixes scale with difficulty. $\Delta$Yes\%: absolute change in yes-rate
(percentage points). $^{**}$\,$p$\,<\,0.01 vs.\ baseline.}
\label{tab:fixbreak}
\end{table}

\paragraph{Key finding: asymmetric fix/break pattern.}
Table~\ref{tab:fixbreak} reveals a striking asymmetry: breaks remain
roughly constant across splits (89--92), while fixes scale with the
number of base errors (88$\to$113$\to$198). On the easiest split,
fixes and breaks nearly cancel (net\,=\,$-1$); on the hardest split,
fixes far exceed breaks (net\,=\,$+106$, $p$\,<\,0.01). This
difficulty dependence is central to our argument: always-on prompting
provides little benefit on easy inputs but yields significant
improvement on hard inputs.

\paragraph{Conservative bias as the underlying mechanism.}
Across all splits, verification prompting consistently reduces the
yes-rate by 5--9 percentage points and shifts the precision--recall
trade-off: Precision increases (e.g., adversarial: 0.747$\to$0.816)
while Recall decreases (0.896$\to$0.836). This conservative shift
helps when the base model over-hallucinates but provides little
benefit when it is already well-calibrated.

\subsection{Attention Redistribution Mechanism}
\label{sec:exp_mechanism}

To understand \emph{how} verification prompting produces the
conservative shift, we extract 32-layer attention maps during prefill
on a subset of POPE-random ($n$\,=\,500) under three conditions:
baseline, verification prompt, and a neutral prompt that adds
non-verification text. We report attention entropy
$H^{(l)}$, visual attention mass $M_\text{vis}^{(l)}$,
and instruction attention mass $M_\text{inst}^{(l)}$ as defined in
Section~\ref{sec:probing}.

\begin{table}[t]
\centering
\small
\begin{tabular}{lrrrrr}
\toprule
\textbf{Layer} & \textbf{$\Delta H_V$\%} & \textbf{$\Delta H_N$\%} & \textbf{Diff} & \textbf{$\Delta M_\text{vis}$} & \textbf{$M_\text{inst}$} \\
\midrule
L1  & $-8.8$  & $-13.3$ & $+4.5$  & $-0.171$ & 0.60 \\
L11 & $+11.2$ & $+1.2$  & $+10.0$ & $-0.025$ & 0.44 \\
L14 & $+2.1$  & $-16.6$ & $+18.7$ & $-0.028$ & 0.40 \\
L15 & $-3.2$  & $-23.0$ & $+19.8$ & $-0.021$ & 0.39 \\
L16 & $-4.4$  & $-20.5$ & $+16.1$ & $-0.040$ & 0.39 \\
L23 & $+6.2$  & $+0.7$  & $+5.6$  & $-0.002$ & 0.15 \\
L31 & $+5.6$  & $+8.4$  & $-2.8$  & $-0.008$ & 0.28 \\
\bottomrule
\end{tabular}
\caption{Attention changes under verification (V) and neutral (N)
prompts relative to baseline (selected layers; full table in Appendix).
$\Delta H$\%: change in attention entropy. Diff =
$\Delta H_V - \Delta H_N$: difference between verification and neutral
controls. $M_\text{inst}$: fraction of attention on instruction tokens
under verification.}
\label{tab:attention}
\end{table}

\paragraph{Finding 1: Instruction tokens absorb attention.}
Under verification prompting, instruction tokens attract 38--60\% of
attention mass in the first 16 layers (Table~\ref{tab:attention},
rightmost column). Visual attention mass decreases in 29 of 32 layers,
with the largest drops in early layers (L1: $-0.171$).

\paragraph{Finding 2: Verification induces a distinct entropy pattern.}
The neutral prompt causes middle-layer entropy to \emph{collapse}
(L14--17: $-17$\% to $-23$\%), whereas verification prompting maintains
or increases entropy in the same layers (L14: $+2.1$\%, L11:
$+11.2$\%). The difference column (Diff) highlights verification-specific
effects that are not explained by input elongation alone: up to $+19.8$\%
at L15.

\paragraph{Interpretation.}
Together, these results suggest that verification prompting changes the
model's internal computation by reallocating attention toward instruction
tokens and maintaining a more dispersed middle-layer attention pattern
than a neutral prompt control. The verification condition does not exhibit
the entropy collapse seen under neutral prompting, and this pattern aligns
with the conservative output shift observed in
Section~\ref{sec:exp_fixbreak}. These observations suggest that
verification prompting is associated with instruction-conditioned
attention redistribution, without clear evidence of uniformly improved
visual grounding.

\subsection{RSP: Selective Intervention}
\label{sec:exp_rsp}

\paragraph{LLaVA-1.5 results.}

\begin{table}[t]
\centering
\small
\begin{tabular}{lcccc}
\toprule
\textbf{Split} & \textbf{Baseline} & \textbf{Always-on} & \textbf{RSP} & \textbf{Trig.\%} \\
\midrule
Random      & .896 & .889          & \textbf{.899} & 7\% \\
Popular     & .864 & .862          & \textbf{.865} & 5\% \\
Adversarial & .810 & \textbf{.827}   & .812          & 6\% \\
\bottomrule
\end{tabular}
\caption{F1 on POPE (LLaVA-1.5). RSP uses L23 attention entropy.
On easy/medium inputs, RSP avoids always-on degradation; on adversarial
inputs, always-on is more beneficial because the fix/break ratio
strongly favors prompting. $^{**}$\,$p$\,<\,0.01.}
\label{tab:main_llava}
\end{table}

Table~\ref{tab:main_llava} shows that RSP remains at or above baseline
on all splits (+0.1--0.3\% F1) while triggering on only 5--7\% of
inputs. On random and popular splits, RSP outperforms always-on
prompting, demonstrating that selective routing avoids the damage caused
by indiscriminate prompting. On adversarial inputs, always-on prompting remains superior
(F1 .827 vs.\ .812 for RSP) because the high base error rate
makes the fix/break balance strongly favor prompting
(Table~\ref{tab:fixbreak}: 198 fixes vs.\ 92 breaks). This does
not contradict RSP's motivation: selective routing is most useful in mixed-difficulty settings, where indiscriminate prompting introduces avoidable breaks on easier inputs. In uniformly hard settings, unconditional prompting can be the better choice.

\paragraph{InstructBLIP results.}

\begin{table}[t]
\centering
\small
\begin{tabular}{lcccc}
\toprule
\textbf{Split} & \textbf{Baseline} & \textbf{Always-on} & \textbf{RSP} & \textbf{Trig.\%} \\
\midrule
Random      & .887 & .874          & \textbf{.891} & 10\% \\
Popular     & .839 & .842          & \textbf{.852} & 14\% \\
Adversarial & .818 & .818          & \textbf{.826} & 14\% \\
\bottomrule
\end{tabular}
\caption{F1 on POPE (InstructBLIP, inverse top-1 confidence,
``Be careful.'' prompt). RSP numerically improves over baseline
on all splits and significantly on popular and adversarial.
$^*$\,$p < 0.05$.}
\label{tab:main_instructblip}
\end{table}

Table~\ref{tab:main_instructblip} shows that RSP with inverse top-1
confidence routing numerically improves over baseline on all three splits
(+0.4--1.3\% F1), with significant gains on popular and adversarial,
and consistently outperforms always-on prompting. The selection signal
must be architecture-aware: middle-layer attention entropy is informative
when visual tokens are directly concatenated (LLaVA), whereas
output-logit confidence is more reliable when visual information is
compressed by a Q-Former (InstructBLIP).

\paragraph{Oracle ceiling and gap analysis.}
An oracle routing analysis (Appendix~\ref{app:oracle}) shows that
perfectly selective prompting on only 3--7\% of inputs could yield
+2.7--5.2\% F1, indicating substantial headroom. The gap between
RSP and oracle performance likely arises because pre-generation
uncertainty captures input-level risk only imperfectly and cannot
reliably predict whether prompting will fix or break a specific
sample. In addition, RSP uses a hard binary threshold, which may
discard borderline cases where adaptive intervention would be
preferable. Closing this gap may require generation-aware signals
or lightweight learned routers, which we leave to future work.

\section{Discussion}
\label{sec:discussion}

RSP currently captures only part of the oracle headroom, indicating
room for stronger routing predictors. The current pre-generation signals
are validated primarily in discriminative settings and do not directly
transfer to open-ended generation, where hallucination may emerge
progressively during decoding. Moreover, effective routing signals
appear architecture-sensitive, and our experiments are limited to
7B-scale LVLMs.

More broadly, our results suggest that verification prompting should
be treated as a risk-bearing intervention rather than a universally
beneficial add-on. Practical systems should decide \emph{when} to
apply verification prompting, not only \emph{how} to formulate it.

\paragraph{Relationship to decoding-based methods.}
Decoding-based hallucination mitigation methods such as
VCD~\citep{leng2024vcd}, OPERA~\citep{huang2024opera}, and
DoLa~\citep{chuang2024dola} modify the generation procedure through
contrastive decoding, attention penalties, or layer-wise logit
contrasting. RSP addresses a different question: given a prompt-based
verification intervention, \emph{when} should it be applied? It is
therefore orthogonal to decoding-time interventions and could in
principle be combined with them. Our goal is not to replace these
methods, but to characterize the risk of uniform prompting and
provide a training-free selective routing mechanism. A full
combination study is beyond the scope of this work.

\section{Conclusion}
\label{sec:conclusion}

We showed that verification prompting in LVLMs is a risk-bearing
intervention: the errors it corrects increase with input difficulty,
while the errors it introduces persist across difficulty levels. This
explains why always-on prompting helps on hard inputs but offers little
benefit---and can harm---easier ones. Our probing analysis shows that
this conservative shift is associated with attention redistribution from
visual tokens toward instruction tokens. Based on these findings, we
proposed RSP, a training-free selective prompting method that uses
pre-generation uncertainty to trigger verification selectively,
mitigating the degradation of indiscriminate prompting while remaining
close to or above baseline performance. Our cross-architecture results
further show that effective routing signals vary across the
architectures we study, motivating future work on stronger and more
general predictors.

\section*{Limitations}

Our work has several limitations. First, our main experiments are
conducted on two open-source LVLMs, LLaVA-1.5 and InstructBLIP. Although
they represent different visual-token interfaces, broader validation on
additional architectures, model scales, and closed-source systems such as
GPT-4V or Gemini remains future work. In particular, some closed-source
models may not expose the internal attention or logit signals required by
RSP.

Second, RSP requires a small development set to select the routing
threshold $\tau$. The optimal threshold may vary across models, datasets,
and task domains, and we do not study zero-shot threshold transfer or
adaptive threshold selection.

Third, RSP is training-free but not inference-free. It requires one probe
prefill pass for each input to extract the uncertainty signal, introducing
additional inference overhead even though verification generation is used
only for the triggered subset.

Fourth, our attention analysis is observational. While we find that
verification prompting is associated with attention redistribution from
visual tokens toward instruction tokens, this does not establish a causal
mechanism or fully explain the model's internal decision process.

Finally, our evaluation focuses primarily on English object-hallucination
settings. Our CHAIR analysis suggests that current pre-generation signals
may not directly transfer to open-ended generation, and multilingual or
cross-lingual multimodal settings remain unexplored.

\bibliography{custom}

\appendix

\section{Prompt Texts}
\label{app:prompts}

\paragraph{LLaVA-1.5 Verification Prompt.}
\begin{quote}
\small\ttfamily
You are a precise visual content describer. Before answering, briefly
verify in your mind: Can I ACTUALLY see this object in the image? Only
answer yes if the object is clearly visible. Do not guess or infer
objects that are not clearly present.
\end{quote}

\paragraph{LLaVA-1.5 Neutral Prompt.}
\begin{quote}
\small\ttfamily
You are a precise visual content describer. RULES: 1) Only describe
objects, attributes, and relationships that are ACTUALLY VISIBLE in the
image. 2) If you are unsure whether something exists in the image, DO
NOT mention it. 3) Do not use your prior knowledge to infer objects that
are not clearly visible. 4) Use hedging language (e.g., `appears to
be', `likely') for anything uncertain.
\end{quote}

\paragraph{InstructBLIP Cautious Prompt.}
\begin{quote}
\small\ttfamily
Be careful.
\end{quote}

\section{Oracle Routing Ceiling}
\label{app:oracle}

\begin{table}[h]
\centering\small
\begin{tabular}{lcccc}
\toprule
\textbf{Split} & \textbf{Baseline} & \textbf{Oracle} & \textbf{$\Delta$F1} & \textbf{Prompt\%} \\
\midrule
Random      & .896 & .924 & +0.028 & 3.0\% \\
Popular     & .864 & .897 & +0.033 & 3.8\% \\
Adversarial & .810 & .862 & +0.052 & 6.6\% \\
\bottomrule
\end{tabular}
\caption{Oracle routing ceiling for LLaVA-1.5. Perfect routing would
prompt only 3--7\% of inputs while yielding substantially larger F1 gains,
indicating headroom for stronger routing signals.}
\label{tab:oracle}
\end{table}

\section{CHAIR Boundary Analysis}
\label{app:chair}

Pre-generation uncertainty signals are near chance for predicting
CHAIR hallucination (prefill entropy AUROC\,=\,0.44), while
generation-time mean entropy provides a moderate signal
(AUROC\,=\,0.63). This contrast suggests that, for open-ended
generation, useful uncertainty information emerges during decoding
rather than being reliably available before generation begins,
representing a boundary of the current pre-generation routing
formulation.

\section{Full Layer-wise Analysis}
\label{app:layer_sweep}

Table~\ref{tab:layer_auroc} reports the AUROC of each layer's
attention entropy for predicting whether verification prompting
will change the model's answer (pilot study, $n$\,=\,100).

\begin{table}[h]
\centering\small
\begin{tabular}{cccccc}
\toprule
\textbf{Layer} & \textbf{AUROC} & \textbf{$\Delta H$\%} & \quad &
\textbf{Layer} & \textbf{AUROC} \\
\midrule
0  & .578 & $-1.8$  && 16 & .526 \\
1  & .528 & $-2.2$  && 17 & .524 \\
2  & .583 & $+16.7$ && 18 & .554 \\
3  & .557 & $+5.1$  && 19 & .504 \\
4  & .580 & $-8.4$  && 20 & .559 \\
5  & .546 & $+2.1$  && 21 & .526 \\
6  & .581 & $+2.9$  && 22 & .616 \\
7  & .536 & $+11.4$ && \textbf{23} & \textbf{.643} \\
8  & .521 & $+9.2$  && 24 & .619 \\
9  & .619 & $+11.4$ && 25 & .558 \\
10 & .509 & $+25.9$ && 26 & .549 \\
11 & .503 & $+7.3$  && 27 & .578 \\
12 & .549 & $+13.6$ && 28 & .575 \\
13 & .529 & $+19.9$ && 29 & .501 \\
14 & .589 & $+28.1$ && 30 & .675 \\
15 & .631 & $+27.2$ && 31 & .516 \\
\bottomrule
\end{tabular}
\caption{Layer-wise AUROC and entropy change under verification
prompting (LLaVA-1.5, pilot $n$\,=\,100). $\Delta H$\%: percentage
change in attention entropy relative to baseline.
Layer~23 is selected on the development set; its full-scale AUROC
is 0.60 (random), 0.67 (popular/adversarial).}
\label{tab:layer_auroc}
\end{table}

\begin{figure}[h]
\centering
\includegraphics[width=\linewidth]{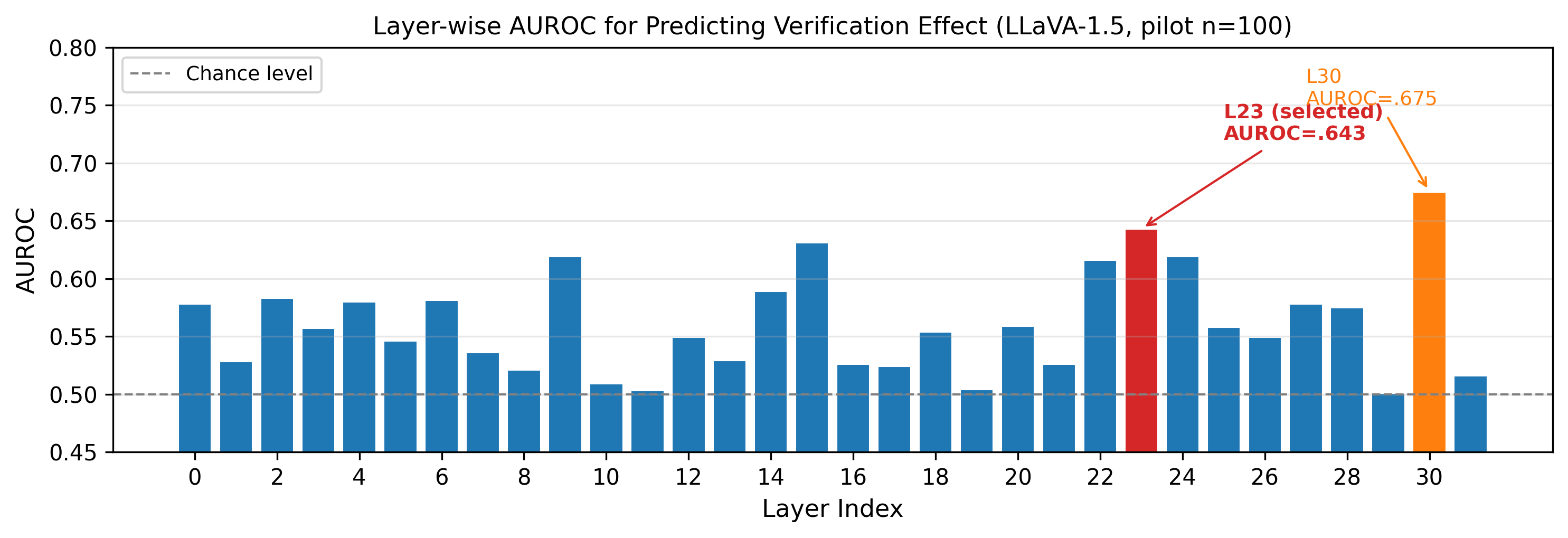}
\caption{Layer-wise AUROC for predicting whether verification
prompting will change the model's answer (LLaVA-1.5, pilot
$n$\,=\,100). Layer~23 (red) is selected; Layer~30 (orange)
achieves higher pilot AUROC but is unstable at larger sample
sizes (Table~\ref{tab:layer_comparison}).}
\label{fig:layer_auroc}
\end{figure}

\paragraph{Why Layer~23 over Layer~30.}
In the pilot study ($n$\,=\,100), Layer~30 achieves the highest
AUROC (.675 vs.\ .643 for Layer~23). However, a larger validation
($n$\,=\,489) reveals that Layer~30's AUROC drops to .298---below
chance and inverted in polarity---while Layer~23's AUROC increases
to .796 (Table~\ref{tab:layer_comparison}). This instability is
also reflected in downstream RSP performance: Layer~30 requires a
50\% trigger rate to achieve its best F1 of .896, whereas Layer~23
achieves F1\,=\,.900 at only 10\% trigger rate. We therefore
select Layer~23 for its stability and efficiency.

\begin{table}[h]
\centering\small
\begin{tabular}{lcccc}
\toprule
\textbf{Layer} & \textbf{Pilot AUROC} & \textbf{Val.\ AUROC}
  & \textbf{Best RSP F1} & \textbf{Trigger\%} \\
 & ($n$\,=\,100) & ($n$\,=\,489) & & \\
\midrule
15 & .631 & .677 & .902 & 30\% \\
\textbf{23} & \textbf{.643} & \textbf{.796}
   & \textbf{.900} & \textbf{10\%} \\
30 & .675 & .298 & .896 & 50\% \\
31 & .516 & .317 & .899 & 50\% \\
\bottomrule
\end{tabular}
\caption{Layer comparison for RSP routing signal (LLaVA-1.5,
POPE-random). Layer~23 has the highest validation AUROC and
achieves competitive F1 at the lowest trigger rate. Layers~30
and~31 show AUROC inversion at larger sample sizes.}
\label{tab:layer_comparison}
\end{table}

\section{Trigger Rate Sensitivity}
\label{app:trigger_rate}

\paragraph{Sensitivity to trigger rate.}
Table~\ref{tab:trigger_ablation} reports RSP performance as the
routing threshold is varied across POPE splits. On random and
popular splits, performance peaks at 5--7\% trigger rate and
degrades as the rate increases toward always-on, confirming that
selective routing at a low trigger rate is beneficial for easy
and medium inputs. On adversarial inputs, performance increases
monotonically with trigger rate, consistent with the finding that
the fix/break ratio strongly favors prompting in high-difficulty
settings (Section~\ref{sec:exp_fixbreak}).

\begin{table}[h]
\centering\small
\begin{tabular}{lccc}
\toprule
\textbf{Trigger rate} & \textbf{Random} & \textbf{Popular}
  & \textbf{Adversarial} \\
\midrule
Baseline (0\%) & .896 & .864 & .810 \\
\midrule
3\%  & .898 & .865 & .810 \\
5\%  & .898 & .864 & .810 \\
7\%$^\dagger$ & \textbf{.899} & \textbf{.865} & .812 \\
10\% & .898 & .865 & .812 \\
15\% & .896 & .864 & .811 \\
20\% & .895 & .863 & .811 \\
50\% & .896 & .865 & .822 \\
\midrule
Always-on (100\%) & .889 & .862 & \textbf{.827} \\
\bottomrule
\end{tabular}
\caption{RSP F1 across trigger rates (LLaVA-1.5, L23 attention
entropy, full POPE splits). $^\dagger$\,=\,threshold used in main experiments. On random and popular splits, low trigger rates outperform always-on; on adversarial, performance increases monotonically with trigger rate.}
\label{tab:trigger_ablation}
\end{table}

\end{document}